\documentclass[letterpaper, 10 pt, conference]{ieeeconf}  % Comment this line out if you need a4paper

\IEEEoverridecommandlockouts                              % This command is only needed if 
\usepackage{graphics} % for pdf, bitmapped graphics files
\usepackage{amsmath} % assumes amsmath package installed
\usepackage{amssymb}  % assumes amsmath package installed
\usepackage{svg}
\usepackage{xcolor}
\usepackage{url}

\usepackage{hyperref}

\title{\LARGE \bf
Open Access NAO (OAN): a ROS2-based software framework for HRI applications with the NAO robot 
}

\author{Antonio Bono$^{a}$, Kenji Brameld$^{b}$, Luigi D'Alfonso$^{a}$ and Giuseppe Fedele$^{a}$% <-this % stops a space
\thanks{*This work was not supported by any organization and it has been submitted to the IEEE for possible publication.  Copyright may be transferred without notice, after which this version may no longer be accessible.}% <-this % stops a space
\thanks{$^{a}$Antonio Bono (corresponding author), Luigi D'Alfonso, Giuseppe Fedele are with Department of Computer Engineering, Modeling, Electronics and Systems,
        University of Calabria, 87036 Rende, Italy.
        {\tt\small \{antonio.bono, luigi.dalfonso, giuseppe.fedele\}@unical.it}}%
\thanks{$^{b}$Kenji Brameld is with TRACLabs Inc.,
16969 North Texas Ave.  
        {\tt\small kenjibrameld@gmail.com}}%
}
\begin{document}
\maketitle
\thispagestyle{empty}
\pagestyle{empty}
%%%%%%%%%%%%%%%%%%%%%%%%%%%%%%%%%%%%%%%%%%%%%%%%%%%%%%%%%%%%%%%%%%%%%%%%%%%%%%%%
\begin{abstract}
This paper presents a new software framework for HRI experimentation with the sixth version of the common NAO robot produced by the United Robotics Group. 
Embracing the common demand of researchers for better performance and new features for NAO, the authors took advantage of the ability to run ROS2 onboard on the NAO to develop a framework independent of the APIs provided by the manufacturer. Such a system provides NAO with not only the basic skills of a humanoid robot such as walking and reproducing movements of interest but also features often used in HRI such as: speech recognition/synthesis, face and object detention, and the use of Generative Pre-trained Transformer (GPT) models for conversation.
The developed code is therefore configured as a ready-to-use but also highly expandable and improvable tool thanks to the possibilities provided by the ROS community.
\end{abstract}
%
%
%%%%%%%%%%%%%%%%%%%%%%%%%%%%%%%%%%%%%%%%%%%%%%%%%%%%%%%%%%%%%%%%%%%%%%%%%%%%%%%%
\section{Introduction}
%
%In recent decades, the efforts of the scientific community in the study of human-robot interaction have greatly intensified, giving rise to a whole field known as human robot interaction (HRI). The many dedicated journals, conferences, and books attempt to cover all the different facets of this topic that is so important for the actual spread of robots in our society: from the study of physical interaction to the study of social relationships with groups of people, from..
%
In the field of Human-Robot Interaction (HRI), the development of methodologies capable of providing the most realistic and satisfactory interaction possible in the various application contexts (education, industry, entertainment, etc.) necessarily passes through substantial experiment campaigns. Indeed, in recent decades, there have been many cases of robots designed and built precisely for the study of interaction with humans: iCub \cite{metta2008icub} is a widespread model within academia for the comprehensive development of cognitive architectures; Kaspar \cite{kaspar} has had excellent results in therapy for Autism Spectrum Disorders (ASDs); Robovie-R4 \cite{robovie} for generic interaction with humans. Some companies have also attempted to disseminate these types of products to a non-specialist audience. Among them, Aldebaran, now part of United Robotics Group, with the development of its first humanoid robot NAO \cite{naodesign} and its successors has certainly established itself as one of the market leaders. 
NAO, in particular, due to its small size, relatively affordable cost, and user-friendly software development environment, has been very successful in both STEM education and HRI research in a wide variety of contexts and with women and men of all ages.
To get an idea of this success, consider that to date more than 15000 units have been sold in 70 countries\footnote{Source: \url{https://unitedrobotics.group/en/robots/nao}}, and since 2008, the year of the first release, several hardware and software upgrades have been made leading up to the sixth and latest version (NAO V6) released in 2018. Scientific productions in which NAO appears from 2010 to 2020 number about 300 according to \cite{naosurvey1}, and several large research projects have been based on this platform, such as: ALIZ-E\footnote{Project website: \url{www.aliz-e.org/}}, CoWriter\footnote{Project website: \url{http://chili.epﬂ.ch/cowriter}}, L2Tor\footnote{Project website: \url{www.l2tor.eu}}. Separate mention deserves its use in the world's largest robotics competition, Robocup, where it has been the selected robot for the Standard Platform League\footnote{Competion webpage: \url{https://spl.robocup.org/}} (SPL) since 2008. Such an impressive body of work and its such heterogeneous use have prompted some authors to even provide survey papers dedicated to it \cite{naosurvey1}\cite{naosurvey2}. These studies cover almost the entire lifespan of this platform and allow for a sense, albeit partial, of how the field of HRI has evolved over the past decade.

As is always the case in applied sciences, methodologies have developed by following and being influenced by the technological evolution of available robotic platforms. The use of increasingly advanced hardware (sensors, processors, etc.) and software tools dedicated to artificial intelligence techniques is widely documented in the literature of the field.
Despite these developments, the available technology has not always lived up to HRI's aspiration toward an interaction experience that is as natural and smooth as possible for the user. NAO is no exception in this regard. Reading the outcomes of the studies analyzed in \cite{naosurvey2}, it is possible to see that hardware/software limitations have often influenced the studies performed in important ways: Tapus et al. \cite{tapus}, in a study of children with ASDs, report how the robot was capable of imitate gross arm movements and it was not fast enough to assure a perfect contingency; in \cite{behrens2018gendered}, a study to investigate how gendered robots voices have influence on a human's trust, the programming opportunities, the ability to modify dialog functions and error states seriously limited the experiments; in \cite{dancetutors}, where NAO plays the role of an instructor to guide the child through several dance moves to learn a dance phrase, the authors note how there are problems with feedback from the robot, that repeated itself or did not adapt to the children; in \cite{sarabia2018assistive} where the robot was teleoperated to be a companion for isolated patients, the authors report a lagged and noisy transmission and the robot was not responsive enough according to the participants; in \cite{pino2020humanoid}, where NAO was used as trainer in a memory program for elderly people with mild cognitive impairment, the researchers conclude the paper stating that it would be interesting if NAO could be totally autonomous once it is switched on and if it could respond to voice command and look like it has its own need.
While, in fact, technological limitations have led to a wide and accepted use of teleoperation in this field, the interest of researchers toward an increasingly realistic autonomy of the robot is demonstrated by an increasing number of scientific papers based on an autonomous mode use of the NAO as reported in \cite{naosurvey1}.

The work presented in this paper aims, precisely, to give an answer to this demand for more performance and autonomy for the NAO platform thanks to a software framework for HRI studies, but not only, in the most popular and modern `operating system' for robotics: ROS2. 
The authors have developed and/or configured a bundle of ROS2 packages that implement some of the basic humanoid robot functionalities: performing user-defined movements (gestures, moves, etc.), walking using feedback to maintain balance, being able to hear and see in the environment thanks to the sensors it is equipped with. 
Keeping well in mind the need to be able to test new algorithms and methods in a protected environment, a simulation software is also provided that allows the user to use the same code that works on the real robot. Through such a tool, we also allow those who do not own the robot to develop highly reusable ROS2 code for the community.
Finally, within this framework, a module has been developed that provides the user with some useful and quite recent features for experimentation on HRI: \textit{i)} recognition of objects or faces in the environment from the cameras and relative tracking by head movement (NAO cannot rotate its eyes); \textit{ii)} communication via voice, thanks to automatic speech recognition and human-like speech synthesis; \textit{iii)} repetition of movements taught kinesthetically by the user (teach by demonstration); \textit{iv)} the ability to interact in numerous languages relying on recent Large Language Models (LLMs), such as Chat-GPT, that allow the identity of the speaking subject to be changed according to the application context. We called such a module Human NAO Interaction (HNI).

In contrast to other software tools that can make Aldebaran's software system communicate with another machine running ROS2, our solution allows \emph{ROS2 to run natively on the robot}.
This possibility positively influences the use of the robot in HRI applications and beyond in at least three respects:
\begin{enumerate}
    \item The elimination of mandatory data transmission with a machine equipped with ROS2, coupled with more efficient control of the hardware, makes it possible to \emph{raise the NAO's hardware performance} to levels hitherto achieved only in the Robocup SPL context.
    \item By using software packages created by the authors, one becomes completely \emph{independent of the closed Aldebaran's API} for high-level robot management. This, in the spirit of open-source code dissemination and participation, will foster collaboration among different developers thus helping the development of increasingly effective algorithms and methods.
    \item The adoption and development of \emph{shared standards and conventions} for ROS software and thus encouraging interoperability and code sharing. 
\end{enumerate}
The desirability of such an open-source approach and the usefulness of common standards is demonstrated by the recent creation of a set of shared conventions and interfaces for HRI formalized in the ROS Enhancement Proposal number 155 (REP-155) \cite{rep155}. This achievement stems precisely from another ROS framework, ROS4HRI, carried out by the authors of \cite{standard}.

To pay homage to the spirit of collaboration and sharing underlying the project, we have titled our framework as: Open Access NAO (OAN). The main open-source code repository can be found here: \url{https://tinyurl.com/oangithub}.

The remainder of the paper is organized as follows: section two provides a summary on the developments of the software dedicated to NAO and the related use of ROS; section three illustrates at a high level the structure of the presented framework; section four details the features for HRI experimentation already implemented; section five presents a simple experiment to show the capabilities of the framework on a real robot; finally, the conclusive section summarizes the main features of the framework and suggests important new improvements and developments.
\section{Background and related works}
The NAO robot, since its first release, has been accompanied by an APIs framework known as NAOqi. This framework, which provides the robot with its interaction capabilities, is accompanied by a graphical development environment, Choreographe \cite{choreograph}, which over the years has been highly regarded for its ease of use especially for in the STEM educational sector. The more advanced applications are, usually, built thanks to two SDKs provided, one for Python 2.x and the other for C++ \cite{nao_sdk}. With this setup, the manufacturer has offered several free applications for using the robot over the years.
The most prominent alternative to such an ecosystem is the Zora Robot Software \cite{zora} from RobotLab Inc, which allows more than 50 different types of activities through the use of a tablet, making it easy to use even for non-specialized personnel such as those in the health care sector. The great success of such software is demonstrated by the numerous publications in which the NAO is presented as a Zora robot like in \cite{van2020zora} and all the references reported in \cite{naosurvey2}.
Both of these solutions stem from the interest of two companies in disseminating the robot to as wide an audience as possible.

The academic robotics community's first interest in NAO as an open platform for research and development dates back to 2009. In that year, Brown University's RLAB released the first ROS software package capable of providing some basic capabilities such as simple navigation in the environment, camera access, text-to-speech, and head control \cite{naobrown}. A few months later, the University of Freiburg augmented those capabilities with a software stack that could provide access to Inertial Measurement Unit (IMU) data, provide for a complete URDF model, and visualize the robot in RViz. The approach of these two pioneering releases was the same: wrapping the needed parts of Aldebaran's NAOqi APIs and making it available in ROS.
This initial project was then developed and maintained over the years independently by Aldebaran through a package that able to make a machine running ROS or ROS2 communicate with the NAO running NAOqiOS (the NAO operating system based on a Linux distribution) \cite{naoqi_driver}.
Such a package, which in fact performs the function of a driver for communication between NAOqiOS and ROS, undoubtedly offers numerous possibilities for interfacing with other software but there are two major drawbacks: the first is the use of only Aldebaran's proprietary APIs for controlling the robot, and the second is the level of performance that is not high given the large amount of data to be exchanged.
In applications where the frequency of interactions and the robot's decision-making speed are not high, such inconveniences can be accepted. When, however, performance demands rise, as in the case of the Robocup soccer competition, they represent a real limitation to the use of the robot itself. As evidence of the less-than-stellar performance of the described solution, consider that no Robocup SPL team makes use of the publicly provided robot management APIs from Aldebaran.
%but they are provided with an ad hoc version of NAOqiOS in such a way that they have direct and higher-performance control of the robot hardware.
In the attempt of overcoming such limits, in 2014, the UChile Robotics Team at Robocup SPL provided a tutorial on how to compile install and run ROS directly on the NAO V4 \cite{forero2014integration}. That capability allowed the team to test in this new setup some algorithms for walking and navigation developed by B-Human, one of the oldest and most successful teams on the SPL.
This effort in the direction of code sharing and diffusion via ROS was unfortunately not followed up during the competition. The reluctance of many teams over the years to publicly release code, in fact, was perceived by the organizing committee as one of the factors that most slowed technical development in the competition, creating a very strong disparity between the various competing teams and making it very difficult for new teams to enter the competition. To mitigate this critical issue, in recent years teams have been obliged to publicly release the code they use, but the lack of common standards still makes interoperability and reuse of different software very difficult.
Sensing this need for common standards, one of the authors, Kenji Brameld, started \textit{ROS Sports} \cite{rossports} an organization that encourages open source collaboration in today's increasingly popular robotic sports competitions. Among the first projects pursued by the author, \texttt{NAO\_LoLA} \cite{naolola} provided the ability to use ROS2 to interface with robot hardware when an ad-hoc version of Ubuntu is installed. Starting from that very capability, the authors developed the ROS framework presented in the remainder of the paper.
\section{The OAN software framework}
In this section we will provide a comprehensive overview of the developed system so that the reader can get an idea of its highly flexible structure and the functionality already implemented. 
\subsection{The underlying operating system}
To control the NAO V6 via ROS2, an Operating System (OS) must be installed on its main processor (Intel Atom® Processor E3845) that can simultaneously interface with all of the robot's peripherals and support the installation of ROS2. The Robocup SPL NAO-Devils team has for some time now been developing and using a customized version of the popular Linux distribution Ubuntu capable of communicating with the robot's sensors and actuators.
Instructions for installing such an OS on the robot are detailed on the dedicated project page \cite{naoimage} and the necessary files are publicly available. With such an OS running on the robot, all that remains is to install the preferred version of ROS2 as reported in the \texttt{NAO\_LoLA} docs \cite{naolola}.
\subsection{The framework components}
\begin{figure}[tp]
	\centering
	\includegraphics[width=0.48\textwidth]{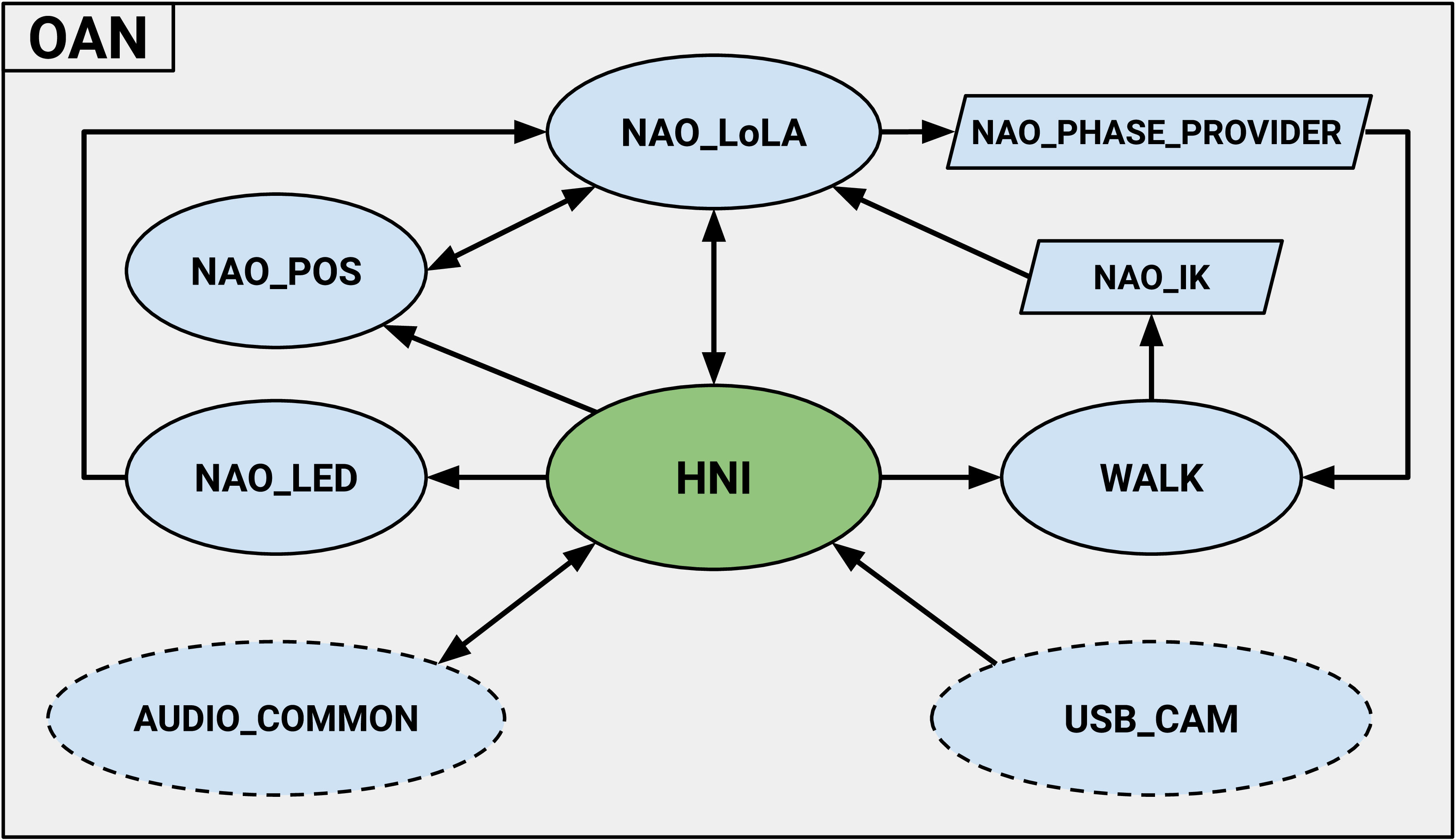}
	\caption{A graph describing the data flow among different parts of the OAN framework. Ellipses and parallelograms are modules and packages, respectively. Dashed lines shows parts developed by the ROS community.}
	\label{fig:scheme}
\end{figure}
The organization of ROS software is based on the release of packages within which are the applications known as nodes. 
%Among the APIs provided for interfacing such applications with the ROS2 Client Library, those written in C++ and Python are certainly the most popular. In our framework we have used both of these APIs giving preference, where possible, to the one written in C++ given the greater possibility of control over the applications and the significantly better performance than the Python-based one. 
Very often, for reasons related to code modularity and the use of third-party software, it is necessary to release not one but a bundle of ROS2 packages in order to implement the desired functionality. Our case is no exception in this regard, and for clarity we will call such bundles as the `modules' of the OAN framework. Figure~\ref{fig:scheme} illustrates their interdependence in a simple way.
The following is a brief description of their main features, except for the one related to the HRI applications, analyzed in more detail in the sequel.
\subsubsection{\texttt{NAO\_LoLA}}
LoLA stands for Low Level Abstraction, a software created by Aldebran to manage the robot hardware. Specifically, this is a real-time process capable of running 83 cycles per second. With each cycle it exchanges data with most of the peripherals. The binaries of such software are embedded within the custom version of Ubuntu released by NAO-Devils. The \texttt{NAO\_LoLA} module allows, precisely, a ROS2 node to be used to interface with such a process. 
%In detail, the module has two main prerogatives: i) receive sensor information from LoLA, split it and publish it to separate ROS2 topics; ii) subscribe to specific ROS2 topics to compose a message for actuators and send it via LoLA.
This makes it possible to receive very important information for robot control, such as those from the IMU, and send messages to the motors at the highest possible frequency (83Hz). This capability, and the associated benefit in robot control, is not achievable using other software solutions that use ROS on an external machine to communicate with NAO.
\subsubsection{\texttt{NAO\_POS}}
The ability to make gestures or generally desired movements is a necessary capability for the development of any realistic interaction with a human user. This module allows the developer to define movements using a file type called `pos' \cite{pos_file}, which has already been widely and successfully used by the SPL Robocup community for several years. In that file each line specifies the position of joints of the robot along with a desired duration for that configuration. When moving from one configuration to the next, a linear interpolation of the joint positions is performed.
The module also allows only certain joints of the robot to be implemented. This capability is critical for simulated execution of multiple applications in motion control (head, arms, hands, etc.) as is normally done in the HRI context.
\subsubsection{\texttt{WALK}}
This module was developed to allow a generic bipedal robot to walk by making use of an algorithm for balance and gait control, which has been used by most Robocup teams for years \cite{hengst}. Just to promote interoperability and reusability of the developed code, there are no specific references to NAO robots in this module. In order to use this module on the NAO V6, as stated in the instructions, it is necessary to use two other packages: \texttt{nao\_phase\_provider} takes care of sensing the pressure with the ground of the NAO's feet thanks to the four available force sensitive resistor mounted on each foot; \texttt{nao\_ik}, on the other hand serves for the calculation of the inverse kinematics of the lower limb and the sending of commands.
Using all these components, it is possible to make the robot walk by giving it direction and speed commands.
%possibly in guided mode as allowed by the \texttt{teleop\_twist\_keyboard} package \cite{teleop}.
%
\subsubsection{\texttt{NAO\_LED}}
This module enables management of the LEDs with which NAO is equipped at a higher level than that provided by \texttt{NAO\_LoLA}. The latter, in fact, allows only simple on/off switching and color selection where possible. In HRI applications, however, light signals can be very useful feedback to the human user, and it is therefore important to provide the ability to create animations with LEDs. Color changing, blinking, possibly cyclic, are some functions offered by the module. These capabilities, which are useful for any robot in the HRI, are even more important for the NAO, which in this context suffers greatly from the inability to change facial expression.
\subsubsection{ROS community modules}
The framework's ability to take advantage of contributions from the ROS community was exploited from the start through the use of two modules developed for generic robots. 
The first, \texttt{audio\_common}, is one of the most widely used ROS modules to interface with audio. Based on the open GStreamer library, this module allows for capturing sounds through the NAO's four microphones and using its two speakers to play sounds.
The second, \texttt{usb\_cam}, is a very common package for interfacing with the robot's cameras. Thanks to the usage of the open V4L library, the module allows streaming of both of the NAO's RGB cameras to ROS2 topics of preference.
Both of these modules are part of the important ROS Drivers project \cite{rosdrivers}, which has been actively maintained by the community since 2011.
\subsection{The related simulator}
\begin{figure}[tbp]
    \centering
    \includegraphics[width=0.4\textwidth]{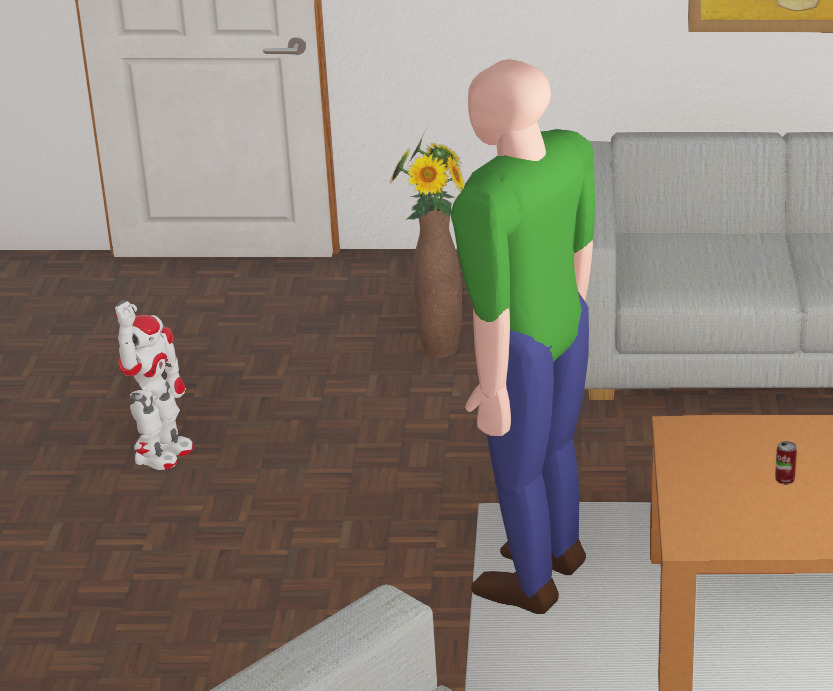}
    \caption{A screenshot from the Webots simulator using the LoLA controller.}
    \label{fig:sim}
\end{figure}

The development of new algorithms and methods for robotics is at least disadvantageous, if not impossible, without using a protected testing environment such as that offered by robot simulation systems such as Gazebo, V-Rep, Webots and others. 
The Robocup SPL teams have been using various solutions for years to enable simulation testing prior to field verification. One in particular, the one developed togheter the Bembelbots team, directly implements the LoLA interface for NAO V6 in a controller for the common Webots simulation system \cite{webots_lola}. It is possible then, to use the ROS package \texttt{NAO\_LoLA} to communicate with such a controller as if it were the real robot. The main advantage of this solution, which is why it was introduced into the framework, is the possibility of \emph{using the same code on both the real and the simulated robot}. Several sensors are also simulated, among which it is worth mentioning the two RGB cameras with which the robot is equipped. Figure~\ref{fig:sim} shows an example of HRI in the simulator. The authors' desire is to allow as soon as possible the use of the audio channel as well so as to have a complete simulation environment for HRI applications in a manner similar to what has been done in \cite{nao_prm}. 
\section{The Human NAO Interaction package}
The Human NAO Interaction (\texttt{HNI}) is a module developed to be able to exploit the potential of the OAN framework in the HRI domain. The goal of this module is, in fact, to provide the user with some very useful capabilities for HRI experimentation, surpassing the current performance allowed by the Aldebaran's ecosystem. As described in the introductory section, the wish of researchers using NAO for better technical capabilities and adoption of new technologies is very strong. The capabilities described below should be considered as a starting point for the development of applications of all sorts through the use of virtually any state-of-the-art library and/or algorithm. These resources are practically usable even on the NAO with limited computational resources thanks to the possibility given by ROS2 to be able to transparently distribute the various applications on several machines in the network.
The possibilities offered by the aforementioned module are outlined below.
\subsubsection{Teach by demonstration}
Defining common movements and gestures for human interaction can be a tedious and time-consuming task. Consider, for example, the definition of a discrete trajectory in joint space to reproduce a gesture of greeting, rather than surprise, etc. The \texttt{joints\_record} node allows us to record the movements of the joints we are interested in. In this way, the user need only move the limbs of interest in the desired manner and the robot will learn the movement. The smooth reproduction of the movement is then handled by the \texttt{joints\_play\_action\_server} node. This solution, based on the teach-by-demonstration approach, is proposed here to allow even nonspecialist users to define new gestures. It is worth mentioning, however, that many other more advanced solutions exist in the literature \cite{speech_gestures_23,speech_gestures_22,speech_gestures_20} and some also specifically designed for NAO \cite{nao_gestures_1}.
\subsubsection{Face/object detection and tracking}
The ability to recognize one or more faces in a real-time video stream is one of the most widely used ways of sensing the presence of a human counterpart. To implement this functionality, we exploited the latest version of the well-known YOLO series of algorithms \cite{yolo}. This newest version, YOLOv8, delivers cutting-edge results for image or video analytics within a straightforward implementation framework. In fact, in order to change the object to be identified, it is only necessary to specify in a single line of code the corresponding pre-trained model. 
Once the object is identified, the \texttt{object\_tracker} node publishes its location inside the RGB frame on a ROS2 topic and therefore the info is available to any other component in the system. 
The \texttt{head\_track\_action\_server} node, in particular, exploits this info to rotate the head of NAO in order to maintaining visual contact with the detected object. This is necessary since NAO cannot move its RGB cameras.
\subsubsection{Speech recognition and synthesis}
The most widely used communication mode in HRI is undoubtedly speech. For this reason, we took care to provide the robot with both the ability to recognize human language and to speak by imitating the human voice. 
These goals were achieved thanks to the the Speech-To-Text (STT) and Text-To-Speech (TTS) APIs offered by the Google Cloud Platform (GCP).  Of the many possibilities that exist, such engines were chosen for at least three reasons: the wide choice of languages and voices to use; robustness to environmental noise; and outstanding performance due to Google's expertise in the development neural network on which they are based. The nodes \texttt{gstt\_service} and \texttt{gtts\_service} are the ones that, practically, manages the GCP services.
%
%The first goal was achieved thanks to the \texttt{gstt\_service} node that uses of one of the most advanced and widespread speech recognition engines in the world: the Speech-To-Text (STT) service offered by Google Cloud Platform (GCP). This tool, based on machine learning, can transcribe text from an audio source in real time, supporting more than 125 languages.
%Model adaptation is employed in STT to enhance the precision of commonly used terms, broaden the transcription vocabulary, and refine transcription quality for audio in noisy environments. This capability coupled with the ability to prioritize recognition of particular words recurring in specific contexts (schools, hospitals, laboratories, etc.), allows the robot recognition at human-like levels.
%Regarding speech synthesis, the \texttt{gtts\_service} node still makes use of the GCP Python APIs, in particular to use the Text-To-Speech (TTS) service. This tool allows for the generation of speech with human intonation while having a choice of more than 380 different voices in more than 50 languages. Based on DeepMind's speech synthesis expertise, it succeeds in reproducing speech with voices that differ little from human voices.
%
\subsubsection{Conversational ability}
The capabilities of the audio channel just exposed, would be little exploitable without being able to establish a conversation with the robot.
NAOqi does not provide a Natural Language Processing (NLP) engine, but only a static database between textual input and output. In fact, no conversation is created but simply a voice interaction with NAO that provides predetermined answers to specific questions given in a list provided to the user \cite{basic_channel}.
Unsatisfied with the Aldebaran's naive mechanism, the community, over the years, has implemented several solutions for NLP on NAO that are gradually more sophisticated \cite{naosurvey1}. %Common factor in all these solutions is the non-negligible implementation effort to integrate the NLP engine with NAOqiOS.
In this package, wanting to take advantage of the recent excellent results provided by Generative Pre-trained Transformer (GPT) models , we used one of the most famous, Chat-GPT, to generate contextualized user responses. Beyond the amazing ability to answer generic questions in a smooth and natural way, Chat-GPT allows for very interesting customization for HRI purposes. 
It is possible, indeed, to modify the personality of the assistant or provide specific instructions about how it should behave throughout the conversation. In our case, therefore, we can have \emph{the model respond as if it were Aldebaran's NAO robot and change its attitudes and preferences}.
An example might be a study in which the model is trained to be particularly talkative and lively in its interaction with children and less expansive and playful with more adult people. Such customization is already possible in the \texttt{chat\_service} node that manages the connection with the model running on the cloud of its producer Open-AI.
\section{A first HRI application}
\begin{figure}[tb]
    \centering
    \includegraphics[width=0.35\textwidth]{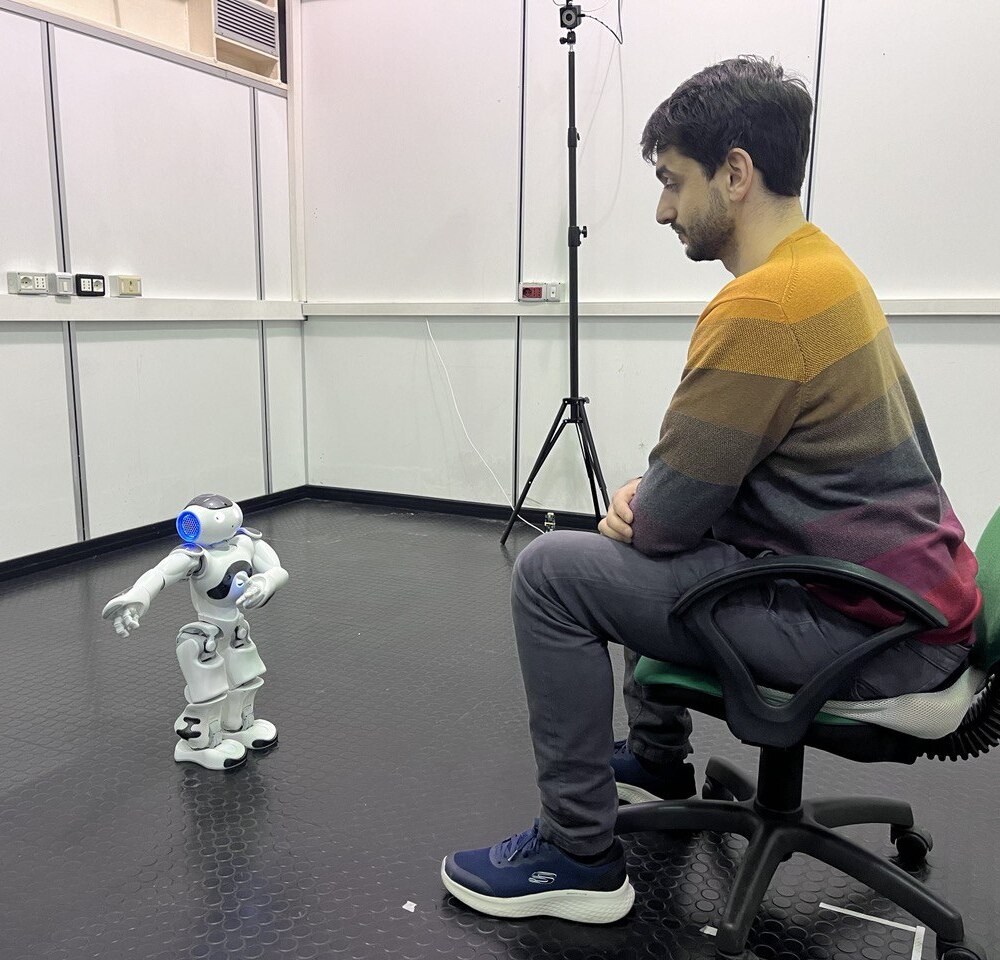}
    \caption{NAO chatting with a user using the OAN framework. The robot is focusing on the user face and plays a gesture while speaking.}
    \label{fig:exp}
\end{figure}
From what has been presented so far, it is clear that the functionalities offered by the OAN framework and, in particular, the HNI module are many. In an attempt to practically show their use and performance, in this section, a simple application of HRI that exploits the presented modules is presented: NAO converses with a human user whose face it recognizes and tracks while reproducing, at the appropriate time, gestures learned through the teach by demonstration approach.
All the modules presented here, with the exception of \texttt{WALK}, were used to implement this behavior. Not mentioning those required for hardware interfacing, we briefly report on their use:
\begin{itemize}
    \item \texttt{NAO\_POS}: to hold the robot upright and make it move in a manner similar to a human moving its legs slightly while talking.
    \item \texttt{NAO\_LED}: to give visual feedback to distinguish various stages of the interaction: listening, processing the response, and playing it back.
    \item \texttt{HNI}: all implemented features were exploited for this application. In particular, the Chat-GPT model is customized to recognize itself as the NAO robot that is used in the DIMES robotics laboratory at the University of Calabria.
\end{itemize} 
The node that orchestrates the use of these functions is \texttt{chat\_action\_server}. The simultaneous execution of the various applications is made possible by two key capabilities: the first is the multi-threading offered by the processor of the NAO V6 and managed through the tools offered by ROS2; the second is the execution of the \texttt{object\_track} node with the YOLOv8 model for face recognition on another networked machine, thus alleviating the computational effort required of the robot. Figure~\ref{fig:exp} shows photo of the interaction, while the full video of the experiment is available at the following link: \url{https://tinyurl.com/hnivideo}.
\section{Conclusions and future developments}
This work stems from the authors' desire to be able to use the very common NAO V6 Robot for state-of-the-art algorithms in HRI. Realizing that such desire for better performance and new features is very common among researchers using it, a new framework was developed for its management totally independent of the high-level control APIs provided by the manufacturer (NAOqi). The cornerstone of the proposed solution is the onboard execution of the most popular and modern robot operating system: ROS2. Taking advantage of the new improved hardware performance and open-source software from the ROS community, we have equipped NAO with some features widely used in HRI experimentation such as: speech recognition/synthesis, face and object detention, and the use of GPT patterns for conversation.
The provided software is, in practice, a starting point for the development and/or testing of new methods for humanoid robotics in the HRI field and beyond. Among the first limitations currently present, which the authors propose to work on, is the use of information from sensors in the GPT model for conversation, thus providing an embodied AI. Another important point certainly remains the implementation of audio sensors within the simulation suite so that more complete interactions with the human user can be analyzed.
Finally, with the intention of adopting shared standards that promote code interoperability, the conventions established in the recent REP-155 for the use of ROS in HRI will be adopted as soon as possible. 
\section*{Acknowledgment}
\addcontentsline{toc}{section}{Acknowledgment} 
The authors thank Aaron Larisch, which is with TU Dortmund (Germany),  for valuable support in understanding and using the custom Ubuntu OS developed by the Robocup NAO-Devils team.
\bibliographystyle{IEEEtran}
\bibliography{root}    

% Generated by IEEEtran.bst, version: 1.14 (2015/08/26)
\begin{thebibliography}{10}
\providecommand{\url}[1]{#1}
\csname url@samestyle\endcsname
\providecommand{\newblock}{\relax}
\providecommand{\bibinfo}[2]{#2}
\providecommand{\BIBentrySTDinterwordspacing}{\spaceskip=0pt\relax}
\providecommand{\BIBentryALTinterwordstretchfactor}{4}
\providecommand{\BIBentryALTinterwordspacing}{\spaceskip=\fontdimen2\font plus
\BIBentryALTinterwordstretchfactor\fontdimen3\font minus \fontdimen4\font\relax}
\providecommand{\BIBforeignlanguage}[2]{{%
\expandafter\ifx\csname l@#1\endcsname\relax
\typeout{** WARNING: IEEEtran.bst: No hyphenation pattern has been}%
\typeout{** loaded for the language `#1'. Using the pattern for}%
\typeout{** the default language instead.}%
\else
\language=\csname l@#1\endcsname
\fi
#2}}
\providecommand{\BIBdecl}{\relax}
\BIBdecl

\bibitem{metta2008icub}
G.~Metta, G.~Sandini, D.~Vernon, L.~Natale, and F.~Nori, ``The icub humanoid robot: an open platform for research in embodied cognition,'' in \emph{Proceedings of the 8th workshop on performance metrics for intelligent systems}, 2008, pp. 50--56.

\bibitem{kaspar}
L.~J. Wood, A.~Zaraki, B.~Robins, and K.~Dautenhahn, ``Developing kaspar: a humanoid robot for children with autism,'' \emph{International Journal of Social Robotics}, vol.~13, no.~3, pp. 491--508, 2021.

\bibitem{robovie}
D.~Das, M.~G. Rashed, Y.~Kobayashi, and Y.~Kuno, ``Supporting human–robot interaction based on the level of visual focus of attention,'' \emph{IEEE Transactions on Human-Machine Systems}, vol.~45, no.~6, pp. 664--675, 2015.

\bibitem{naodesign}
D.~Gouaillier, V.~Hugel, P.~Blazevic, C.~Kilner, J.~Monceaux, P.~Lafourcade, B.~Marnier, J.~Serre, and B.~Maisonnier, ``Mechatronic design of nao humanoid,'' in \emph{2009 IEEE International Conference on Robotics and Automation}, 2009, pp. 769--774.

\bibitem{naosurvey1}
A.~Amirova, N.~Rakhymbayeva, E.~Yadollahi, A.~Sandygulova, and W.~Johal, ``10 years of human-nao interaction research: A scoping review,'' \emph{Frontiers in Robotics and AI}, vol.~8, 2021.

\bibitem{naosurvey2}
A.~Robaczewski, J.~Bouchard, K.~Bouchard, and S.~Gaboury, ``Socially assistive robots: The specific case of the nao,'' \emph{International Journal of Social Robotics}, vol.~13, pp. 795--831, 2021.

\bibitem{tapus}
\BIBentryALTinterwordspacing
A.~Tapus, A.~Peca, A.~Aly, C.~Pop, L.~Jisa, S.~Pintea, A.~S. Rusu, and D.~O. David, ``Children with autism social engagement in interaction with nao, an imitative robot: A series of single case experiments,'' \emph{Interaction Studies}, vol.~13, no.~3, pp. 315--347, 2012. [Online]. Available: \url{https://www.jbe-platform.com/content/journals/10.1075/is.13.3.01tap}
\BIBentrySTDinterwordspacing

\bibitem{behrens2018gendered}
S.~I. Behrens, A.~K.~K. Egsvang, M.~Hansen, and A.~M. M{\o}lleg{\aa}rd-Schroll, ``Gendered robot voices and their influence on trust,'' in \emph{Companion of the 2018 ACM/IEEE international conference on human-robot interaction}, 2018, pp. 63--64.

\bibitem{dancetutors}
\BIBentryALTinterwordspacing
R.~Ros, I.~Baroni, and Y.~Demiris, ``Adaptive human–robot interaction in sensorimotor task instruction: From human to robot dance tutors,'' \emph{Robotics and Autonomous Systems}, vol.~62, no.~6, pp. 707--720, 2014. [Online]. Available: \url{https://www.sciencedirect.com/science/article/pii/S0921889014000499}
\BIBentrySTDinterwordspacing

\bibitem{sarabia2018assistive}
M.~Sarabia, N.~Young, K.~Canavan, T.~Edginton, Y.~Demiris, and M.~P. Vizcaychipi, ``Assistive robotic technology to combat social isolation in acute hospital settings,'' \emph{International Journal of Social Robotics}, vol.~10, pp. 607--620, 2018.

\bibitem{pino2020humanoid}
O.~Pino, G.~Palestra, R.~Trevino, and B.~De~Carolis, ``The humanoid robot nao as trainer in a memory program for elderly people with mild cognitive impairment,'' \emph{International Journal of Social Robotics}, vol.~12, pp. 21--33, 2020.

\bibitem{rep155}
S.~Lemaignan, ``{ROS} {E}nhancment {P}roposal number 155,'' \url{https://ros.org/reps/rep-0155.html}, accessed: (2024-03-15).

\bibitem{standard}
Y.~Mohamed and S.~Lemaignan, ``Ros for human-robot interaction,'' in \emph{2021 IEEE/RSJ International Conference on Intelligent Robots and Systems (IROS)}, 2021, pp. 3020--3027.

\bibitem{choreograph}
E.~Pot, J.~Monceaux, R.~Gelin, and B.~Maisonnier, ``Choregraphe: a graphical tool for humanoid robot programming,'' in \emph{RO-MAN 2009 - The 18th IEEE International Symposium on Robot and Human Interactive Communication}, 2009, pp. 46--51.

\bibitem{nao_sdk}
{Softbanks Robotics}, ``{NAOqi SDKs},'' \url{http://doc.aldebaran.com/2-5/dev/programming_index.html}, accessed: (2024-03-15).

\bibitem{zora}
{RobotLAB}, ``Zora robot software,'' \url{https://www.robotlab.com/store/zora-robot-solution-for-healthcare}, accessed: (2024-03-15).

\bibitem{van2020zora}
R.~J. van~den Heuvel, M.~A. Lexis, and L.~P. de~Witte, ``Zora robot based interventions to achieve therapeutic and educational goals in children with severe physical disabilities,'' \emph{International Journal of Social Robotics}, vol.~12, no.~2, pp. 493--504, 2020.

\bibitem{naobrown}
K.~Conley, ``Robots using {ROS}: Aldebaran {NAO},'' \url{https://www.ros.org/news/2010/03/robots-using-ros-aldebaran-nao.html}, accessed: (2024-03-15).

\bibitem{naoqi_driver}
{Aldebaran}, ``{ROS2} driver for {NAO} and {Pepper} robots,'' \url{https://index.ros.org/p/naoqi_driver/}, accessed: (2024-03-15).

\bibitem{forero2014integration}
L.~L. Forero, J.~M. Y{\'a}nez, and J.~Ruiz-del Solar, ``Integration of the {ROS} framework in soccer robotics: the {NAO} case,'' in \emph{RoboCup 2013: Robot World Cup XVII 17}.\hskip 1em plus 0.5em minus 0.4em\relax Springer, 2014, pp. 664--671.

\bibitem{rossports}
K.~Brameld, ``Ros {S}ports official documentation,'' \url{https://ros-sports.readthedocs.io/en/latest/index.html}, 2022, accessed: (2024-03-15).

\bibitem{naolola}
------, ``{NAO} {LoLA} official documentation,'' \url{https://nao-lola.readthedocs.io/en/rolling/index.html}, accessed: (2024-03-15).

\bibitem{naoimage}
{Nao-Devils SPL Robocup Team}, ``Nao {I}mage,'' \url{https://github.com/NaoDevils/NaoImage}, accessed: (2024-03-15).

\bibitem{pos_file}
{rUNSWift Robocup SPL Team}, ``{.pos files documentation},'' \url{https://runswift.readthedocs.io/en/latest/motion/index.html#pos-files}, accessed: (2024-03-15).

\bibitem{hengst}
B.~Hengst, M.~Lange, and B.~White, ``Learning ankle-tilt and foot-placement control for flat-footed bipedal balancing and walking,'' in \emph{2011 11th IEEE-RAS International Conference on Humanoid Robots}, 2011, pp. 288--293.

\bibitem{rosdrivers}
{ROS Special Interest Group for drivers}, ``{ROS} drivers,'' \url{https://wiki.ros.org/sig/Drivers}, accessed: (2024-03-15).

\bibitem{webots_lola}
{Bembelbots SPL Team}, ``Webots {LoLA} controller,'' \url{https://github.com/Bembelbots/WebotsLoLaController}, accessed: (2024-03-15).

\bibitem{nao_prm}
C.~Gena, C.~Mattutino, W.~Maltese, G.~Piazza, and E.~Rizzello, ``{Nao\_PRM}: an interactive and affective simulator of the nao robot,'' in \emph{2021 30th IEEE International Conference on Robot and Human Interactive Communication (RO-MAN)}, 2021, pp. 727--734.

\bibitem{speech_gestures_23}
C.~Y. Liu, G.~Mohammadi, Y.~Song, and W.~Johal, ``Speech-gesture gan: Gesture generation for robots and embodied agents,'' in \emph{2023 32nd IEEE International Conference on Robot and Human Interactive Communication (RO-MAN)}, 2023, pp. 405--412.

\bibitem{speech_gestures_22}
U.~Zabala, I.~Rodriguez, and E.~Lazkano, ``Towards an automatic generation of natural gestures for a storyteller robot,'' in \emph{2022 31st IEEE International Conference on Robot and Human Interactive Communication (RO-MAN)}, 2022, pp. 1209--1215.

\bibitem{speech_gestures_20}
N.~T. Viet~Tuyen, A.~Elibol, and N.~Y. Chong, ``Conditional generative adversarial network for generating communicative robot gestures,'' in \emph{2020 29th IEEE International Conference on Robot and Human Interactive Communication (RO-MAN)}, 2020, pp. 201--207.

\bibitem{nao_gestures_1}
M.~Hellou, N.~Gasteiger, A.~Kweon, J.~Lim, B.~A. MacDonald, A.~Cangelosi, and H.~S. Ahn, ``Development and validation of a motion dictionary to create emotional gestures for the nao robot,'' in \emph{2023 32nd IEEE International Conference on Robot and Human Interactive Communication (RO-MAN)}, 2023, pp. 897--902.

\bibitem{yolo}
P.~Jiang, D.~Ergu, F.~Liu, Y.~Cai, and B.~Ma, ``A review of {YOLO} algorithm developments,'' \emph{Procedia computer science}, vol. 199, pp. 1066--1073, 2022.

\bibitem{basic_channel}
{Softbanks Robotics}, ``{Basic Channel - What can I say to NAO},'' \url{http://doc.aldebaran.com/2-8/family/nao_user_guide/spanish.html}, accessed: (2024-03-15).

\end{thebibliography}

\addtolength{\textheight}{-12cm}   % This command serves to balance the column lengths
                                  % on the last page of the document manually. It shortens
                                  % the textheight of the last page by a suitable amount.
                                  % This command does not take effect until the next page
                                  % so it should come on the page before the last. Make
                                  % sure that you do not shorten the textheight too much.

\end{document}